\documentclass[11pt,letterpaper]{article}
\usepackage{setspace}
\usepackage{naaclhlt2016}
\usepackage{times}
\usepackage{latexsym}
\usepackage{graphicx}
\usepackage{color}
\usepackage{multirow}
\usepackage{amsfonts}
\usepackage{amsmath}
\usepackage{tablefootnote}
\usepackage{pbox}
\usepackage{float}
\usepackage{adjustbox}
\usepackage{array}

\naaclfinalcopy 
\def\naaclpaperid{***} 

\newcolumntype{R}[2]{%
    >{\adjustbox{angle=#1,lap=\width-(#2)}\bgroup}%
    l%
    <{\egroup}%
}

\newcounter{note-counter}

\title{Capturing Semantic Similarity for Entity Linking \\ with Convolutional Neural Networks}

\author{Matthew Francis-Landau,\ \ Greg Durrett \and Dan Klein \\
  Computer Science Division\\
  University of California, Berkeley\\
  {\tt \{mfl,gdurrett,klein\}@cs.berkeley.edu}
  }

\date{}

\begin{document}

\maketitle

\begin{abstract}
A key challenge in entity linking is making effective use of contextual information to disambiguate mentions that might refer to different entities in different contexts. We present a model that uses convolutional neural networks to capture semantic correspondence between a mention's context and a proposed target entity. These convolutional networks operate at multiple granularities to exploit various kinds of topic information, and their rich parameterization gives them the capacity to learn which $n$-grams characterize different topics. We combine these networks with a sparse linear model to achieve state-of-the-art performance on multiple entity linking datasets, outperforming the prior systems of \newcite{DurrettJoint2014} and \newcite{NguyenAidaLight2015}.\footnote{Source available at \\ \texttt{github.com/matthewfl/nlp-entity-convnet}}
\end{abstract}

\section{Introduction}
One of the major challenges of entity linking is resolving contextually polysemous mentions. For example, \emph{Germany} may refer to a nation, to that nation's government, or even to a soccer team. Past approaches to such cases have often focused on collective entity linking: nearby mentions in a document might be expected to link to topically-similar entities, which can give us clues about the identity of the mention currently being resolved \cite{RatinovWiki2011,HoffartCoNLL2011,HeStacking2013,ChengRoth2013,DurrettJoint2014}. But an even simpler approach is to use context information from just the words in the source document itself to make sure the entity is being resolved sensibly in context. In past work, these approaches have typically relied on heuristics such as tf-idf \cite{RatinovWiki2011}, but such heuristics are hard to calibrate and they capture structure in a coarser way than learning-based methods.

In this work, we model semantic similarity between a mention's source document context and its potential entity targets using convolutional neural networks (CNNs). CNNs have been shown to be effective for sentence classification tasks \cite{KalchbrennerEtAl2014,KimCNNSent2014,IyyerEtAl2015} and for capturing similarity in models for entity linking \cite{SunNNetEntity2015} and other related tasks \cite{DongEtAl2015,ShenEtAl2014}, so we expect them to be effective at isolating the relevant topic semantics for entity linking. We show that convolutions over multiple granularities of the input document are useful for providing different notions of semantic context. Finally, we show how to integrate these networks with a preexisting entity linking system \cite{DurrettJoint2014}. Through a combination of these two distinct methods into a single system that leverages their complementary strengths, we achieve state-of-the-art performance across several datasets.

\section{Model}
\begin{figure*}[ht]
  \centering
  \includegraphics[width=\textwidth]{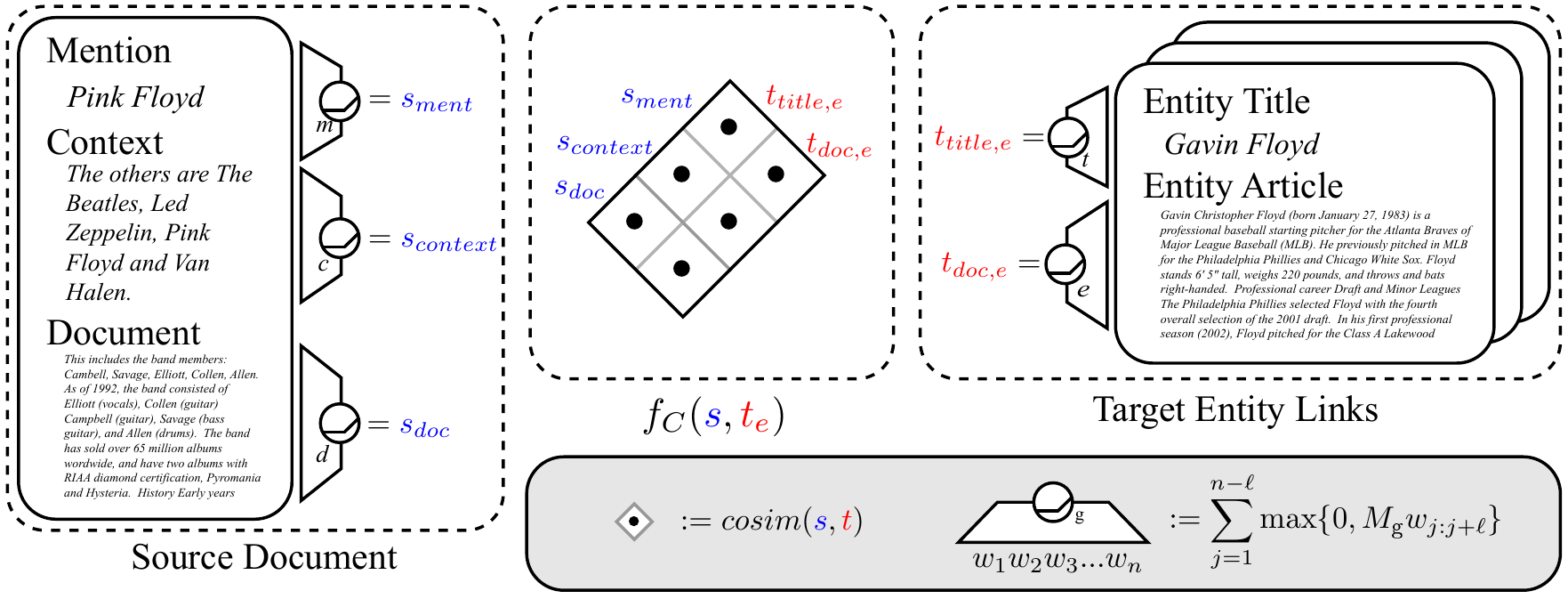}
  \caption{Extraction of convolutional vector space features $f_C(x,t_e)$. Three types of information from the input document and two types of information from the proposed title are fed through convolutional networks to produce vectors, which are systematically compared with cosine similarity to derive real-valued semantic similarity features.}
  \label{fig:conv1}
\end{figure*}

Our model focuses on two core ideas: first, that topic semantics at different granularities in a document are helpful in determining the genres of entities for entity linking, and second, that CNNs can distill a block of text into a meaningful topic vector.

Our entity linking model is a log-linear model that places distributions over target entities $t$ given a mention $x$ and its containing source document. For now, we take $P(t|x) \propto \exp w^\top f_C(x, t; \theta)$, where $f_C$ produces a vector of features based on CNNs with parameters $\theta$ as discussed in Section~\ref{sec:dense}. Section~\ref{sec:sparse} describes how we combine this simple model with a full-fledged entity linking system. As shown in the middle of Figure~\ref{fig:conv1}, each feature in $f_C$ is a cosine similarity between a topic vector associated with the source document and a topic vector associated with the target entity. These vectors are computed by distinct CNNs operating over different subsets of relevant text.

Figure~\ref{fig:conv1} shows an example of why different kinds of context are important for entity linking. In this case, we are considering whether \emph{Pink Floyd} might link to the article \texttt{Gavin\_Floyd} on Wikipedia (imagine that \emph{Pink Floyd} might be a person's nickname). If we look at the source document, we see that the immediate source document context around the mention \emph{Pink Floyd} is referring to rock groups (\emph{Led Zeppelin}, \emph{Van Halen}) and the target entity's Wikipedia page is primarily about sports (\emph{baseball starting pitcher}). Distilling these texts into succinct topic descriptors and then comparing those helps tell us that this is an improbable entity link pair. In this case, the broader source document context actually does not help very much, since it contains other generic last names like \emph{Campbell} and \emph{Savage} that might not necessarily indicate the document to be in the music genre. However, in general, the whole document might provide a more robust topic estimate than a small context window does.

\subsection{Convolutional Semantic Similarity} \label{sec:dense}

Figure~\ref{fig:conv1} shows our method for computing topic vectors and using those to extract features for a potential Wikipedia link. For each of three text granularities in the source document (the mention, that mention's immediate context, and the entire document) and two text granularities on the target entity side (title and Wikipedia article text), we produce vector representations with CNNs as follows. We first embed each word into a $d$-dimensional vector space using standard embedding techniques (discussed in Section~\ref{sec:embeddingVecs}), yielding a sequence of vectors $w_1,\ldots,w_n$. We then map those words into a fixed-size vector using a convolutional network parameterized with a filter bank $M \in \mathbb{R}^{k \times d\ell}$. We put the result through a rectified linear unit (ReLU) and combine the results with sum pooling, giving the following formulation:

\begin{equation} \label{eqn:conv}
  \mbox{conv}_\textrm{g}(w_{1:n}) = \sum_{j=1}^{n-\ell} \max \{ 0, M_\textrm{g} w_{j:j+\ell} \}
\end{equation}
where $w_{j:j+\ell}$ is a concatenation of the given word vectors and the max is element-wise.\footnote{For all experiments, we set $\ell = 5$ and $k = 150$.} Each convolution granularity (mention, context, etc.) has a distinct set of filter parameters $M_\textrm{g}$.

This process produces multiple representative topic vectors $s_{ment}$, $s_{context}$, and $s_{doc}$ for the source document and $t_{title}$ and $t_{doc}$ for the target entity, as shown in Figure \ref{fig:conv1}. All pairs of these vectors between the source and the target are then compared using cosine similarity, as shown in the middle of Figure~\ref{fig:conv1}. This yields the vector of features $f_C(s,t_e)$ which indicate the different types of similarity; this vector can then be combined with other sparse features and fed into a final logistic regression layer (maintaining end-to-end inference and learning of the filters). When trained with backpropagation, the convolutional networks should learn to map text into vector spaces that are informative about whether the document and entity are related or not.

\subsection{Integrating with a Sparse Model} \label{sec:sparse}

The dense model presented in Section~\ref{sec:dense} is effective at capturing semantic topic similarity, but it is most effective when combined with other signals for entity linking. An important cue for resolving a mention is the use of link counts from hyperlinks in Wikipedia \cite{Cucerzan2007,MilneWitten2008,JiGrishman2011}, which tell us how often a given mention was linked to each article on Wikipedia. This information can serve as a useful prior, but only if we can leverage it effectively by targeting the most salient part of a mention. For example, we may have never observed \emph{President Barack Obama} as a linked string on Wikipedia, even though we have seen the substring \emph{Barack Obama} and it unambiguously indicates the correct answer.

Following \newcite{DurrettJoint2014}, we introduce a latent variable $q$ to capture which subset of a mention (known as a \emph{query}) we resolve. Query generation includes potentially removing stop words, plural suffixes, punctuation, and leading or tailing words.  This processes generates on average 9 queries for each mention. Conveniently, this set of queries also defines the set of candidate entities that we consider linking a mention to: each query generates a set of potential entities based on link counts, whose unions are then taken to give on the possible entity targets for each mention (including the null link). In the example shown in Figure~\ref{fig:conv1}, the query phrases are \emph{Pink Floyd} and \emph{Floyd}, which generate \texttt{Pink\_Floyd} and \texttt{Gavin\_Floyd} as potential link targets (among other options that might be derived from the \emph{Floyd} query).

Our final model has the form $P(t|x) = \sum_q P(t,q|x)$. We parameterize $P(t,q|x)$ in a log-linear way with three separate components:
\begin{equation*}
\resizebox{1.0\hsize}{!}{$P(t,q|x) \propto \exp\left( w^\top (f_Q(x, q) + f_E(x, q, t) + f_C(x, t;\theta)) \right)$}
\end{equation*}
$f_Q$ and $f_E$ are both sparse features vectors and are taken from previous work \cite{DurrettJoint2014}. $f_C$ is as discussed in Section~\ref{sec:dense}. Note that $f_C$ has its own internal parameters $\theta$ because it relies on CNNs with learned filters; however, we can compute gradients for these parameters with standard backpropagation. The whole model is trained to maximize the log likelihood of a labeled training corpus using Adadelta \cite{ZeilerAdadelta2012}.

The indicator features $f_Q$ and $f_E$ are described in more detail in \newcite{DurrettJoint2014}. $f_Q$ only impacts which query is selected and not the disambiguation to a title. It is designed to roughly capture the basic shape of a query to measure its desirability, indicating whether suffixes were removed and whether the query captures the capitalized subsequence of a mention, as well as standard lexical, POS, and named entity type features. $f_E$ mostly captures how likely the selected query is to correspond to a given entity based on factors like anchor text counts from Wikipedia, string match with proposed Wikipedia titles, and discretized cosine similarities of tf-idf vectors \cite{RatinovWiki2011}. Adding tf-idf indicators is the only modification we made to the features of \newcite{DurrettJoint2014}.

\section{Experimental Results}
\label{sec:experiments}

We performed experiments on 4 different entity linking datasets.

\begin{itemize}
\item ACE \cite{Ace2005,BentivogliEtAl2010}: This corpus was used in \newcite{FahrniLatent2014} and \newcite{DurrettJoint2014}.
\item CoNLL-YAGO \cite{HoffartCoNLL2011}: This corpus is based on the CoNLL 2003 dataset; the test set consists of 231 news articles and contains a number of rarer entities.
\item WP \cite{HeathWPLink2011}: This dataset consists of short snippets from Wikipedia.
\item Wikipedia \cite{RatinovWiki2011}: This dataset consists of 10,000 randomly sampled Wikipedia articles, with the task being to resolve the links in each article.\footnote{We do not compare to \newcite{RatinovWiki2011} on this dataset because we do not have access to the original Wikipedia dump they used for their work and as a result could not duplicate their results or conduct comparable experiments, a problem which was also noted by \newcite{NguyenAidaLight2015}.}
\end{itemize}
We use standard train-test splits for all datasets except for WP, where no standard split is available. In this case, we randomly sample a test set. For all experiments, we use word vectors computed by running word2vec \cite{MikolovW2V2013} on all Wikipedia, as described in Section \ref{sec:embeddingVecs}.

\begin{table*}[ht]
  \small
  \centering
\begin{tabular}{|l|l|l|}
  \hline
  destroying missiles . spy planes            & has died of his wounds              & him which was more depressing          \\
  and destroying missiles . spy               & vittorio sacerdoti has told his     & a trip and you see                     \\
  by U.N. weapons inspectors .                & his bail hearing , his              & `` i can see why                       \\
  inspectors are discovering and destroying   & bail hearing , his lawyer           & i think so many americans              \\
  are discovering and destroying missiles     & died of his wounds after            & his life from the depression           \\
  an attack using chemical weapons            & from scott peterson 's attorney     & trip and you see him                   \\
  discovering and destroying missiles .       & 's murder trial . she               & , but dumb liberal could               \\
  attack munitions or j-dam weapons           & has told his remarkable tale        & i can see why he                       \\
  sanctions targeting iraq civilians ,        & murder trial . she asking           & one passage . you cite                 \\
  its nuclear weapons and missile             & trial lawyers are driving doctors   & think so many americans are            \\
  \hline
\end{tabular}
\caption{Five-grams representing the maximal activations from different filters in the convolution over the source document ($M_{doc}$, producing $s_{doc}$ in Figure~\ref{fig:conv1}).  Some filters tend towards singular topics as shown in the first and second columns, which focus on weapons and trials, respectively. Others may have a mix of seemingly unrelated topics, as shown in the third column, which does not have a coherent theme. However, such filters might represent a superposition of filters for various topics which never cooccur and thus never need to be disambiguated between.}
\label{table:argmax}
\end{table*}

\begin{table}[t]
  \centering
  \small
  \begin{tabular}{|l|cccc|} \hline
      & ACE & CoNLL & WP & Wiki\tablefootnote{The test set for this dataset is only 40 out of 10,000 documents and subject to wide variation in performance.} \\ \hline
 \multicolumn{5}{|c|}{Previous work} \\ \hline
 DK2014        & 79.6 & ---  & --- & ---  \\
 AIDA-LIGHT     &   --- & 84.8  & ---   & --- \\ \hline
  \multicolumn{5}{|c|}{This work} \\ \hline
 Sparse features & 83.6 & 74.9 & 81.1 & 81.5 \\
 CNN features     & 84.5 & 81.2 & 87.7 & 75.7
 \\
 Full           & \textbf{89.9} & \textbf{85.5} & \textbf{90.7}  & \textbf{82.2} \\ \hline
  \end{tabular}
  \caption{Performance of the system in this work (Full) compared to two baselines from prior work and two ablations. Our results outperform those of \protect\newcite{DurrettJoint2014} and \protect\newcite{NguyenAidaLight2015}. In general, we also see that the convolutional networks by themselves can outperform the system using only sparse features, and in all cases these stack to give substantial benefit.}
  \label{table:resultsAll}
\end{table}

Table~\ref{table:resultsAll} shows results for two baselines and three variants of our system. Our main contribution is the combination of indicator features and CNN features (Full). We see that this system outperforms the results of \newcite{DurrettJoint2014} and the AIDA-LIGHT system of \newcite{NguyenAidaLight2015}. We can also compare to two ablations: using just the sparse features (a system which is a direct extension of \newcite{DurrettJoint2014}) or using just the CNN-derived features.\footnote{In this model, the set of possible link targets for each mention is still populated using anchor text information from Wikipedia (Section~\ref{sec:sparse}), but note that link counts are not used as a feature here.} Our CNN features generally outperform the sparse features and improve even further when stacked with them. This reflects that they capture orthogonal sources of information: for example, the sparse features can capture how frequently the target document was linked to, whereas the CNNs can capture document context in a more nuanced way.  These CNN features also clearly supersede the sparse features based on tf-idf (taken from \cite{RatinovWiki2011}), showing that indeed that CNNs are better at learning semantic topic similarity than heuristics like tf-idf.

In the sparse feature system, the highest weighted features are typically those indicating the frequency that a page was linked to and those indicating specific lexical items in the choice of the latent query variable $q$. This suggests that the system of \newcite{DurrettJoint2014} has the power to pick the right span of a mention to resolve, but then is left to generally pick the most common link target in Wikipedia, which is not always correct. By contrast, the full system has a greater ability to pick less common link targets if the topic indicators distilled from the CNNs indicate that it should do so.

\begin{table}[t!]
  \centering
  \begin{tabular}{|l|ccc|} \hline
  & ACE & CoNLL & WP \\ \hline
  $cosim(s_{doc},t_{doc})$ & 77.43 & 79.76 & 72.93 \\
  $cosim(s_{ment},t_{title})$ & 80.19 & 80.86  & 70.25 \\
  All CNN pairs & 84.85 & 86.91 & 82.02 \\ \hline
  \end{tabular}
  \caption{Comparison of using only topic information derived from the document and target article, only information derived from the mention itself and the target entity title, and the full set of information (six features, as shown in Figure~\ref{fig:conv1}). Neither the finest nor coarsest convolutional context can give the performance of the complete set. Numbers are reported on a development set.}
  \label{table:SimpleVAllConv}
\end{table}

\subsection{Multiple Granularities of Convolution}

One question we might ask is how much we gain by having multiple convolutions on the source and target side. Table \ref{table:SimpleVAllConv} compares our full suite of CNN features, i.e.~the six features specified in Figure~\ref{fig:conv1}, with two specific convolutional features in isolation. Using convolutions over just the source document ($s_{doc}$) and target article text ($t_{doc}$) gives a system\footnote{This model is roughly comparable to Model 2 as presented in \newcite{SunNNetEntity2015}.} that performs, in aggregate, comparably to using convolutions over just the mention ($s_{ment}$) and the entity title ($t_{title}$). These represent two extremes of the system: consuming the maximum amount of context, which might give the most robust representation of topic semantics, and consuming the minimum amount of context, which gives the most focused representation of topics semantics (and which more generally might allow the system to directly memorize train-test pairs observed in training). However, neither performs as well as the combination of all CNN features, showing that the different granularities capture complementary aspects of the entity linking task.


\begin{table}[ht]
  \centering
  \begin{tabular}{|l|ccc|}
	\hline
   & ACE & CoNLL & WP \\ \hline
   Google News & 87.5 & 89.6 & 83.8 \\
   Wikipedia & 89.5 & 90.6 & 85.5 \\ \hline
   \end{tabular}
  \caption{Results of the full model (sparse and convolutional features) comparing word vectors derived from Google News vs. Wikipedia on development sets for each corpus.}
  \label{table:gnewsvecs}
\end{table}

\subsection{Embedding Vectors} \label{sec:embeddingVecs}

We also explored two different sources of embedding vectors for the convolutions. Table \ref{table:gnewsvecs} shows that word vectors trained on Wikipedia outperformed Google News word vectors trained on a larger corpus.  Further investigation revealed that the Google News vectors had much higher out-of-vocabulary rates.  For learning the vectors, we use the standard word2vec toolkit \cite{MikolovW2V2013} with vector length set to 300, window set to 21 (larger windows produce more semantically-focused vectors \cite{LevyGoldberg2014}), 10 negative samples and 10 iterations through Wikipedia. We do not fine-tune word vectors during training of our model, as that was not found to improve performance.

\subsection{Analysis of Learned Convolutions}

One downside of our system compared to its purely indicator-based variant is that its operation is less interpretable. However, one way we can inspect the learned system is by examining what causes high activations of the various convolutional filters (rows of the matrices $M_g$ from Equation~\ref{eqn:conv}). Table~\ref{table:argmax} shows the $n$-grams in the ACE dataset leading to maximal activations of three of the filters from $M_{doc}$. Some filters tend to learn to pick up on $n$-grams characteristic of a particular topic.  In other cases, a single filter might be somewhat inscrutable, as with the third column of Table~\ref{table:argmax}. There are a few possible explanations for this. First, the filter may generally have low activations and therefore have little impact in the final feature computation. Second, the extreme points of the filter may not be characteristic of its overall behavior, since the bulk of $n$-grams will lead to more moderate activations. Finally, such a filter may represent the superposition of a few topics that we are unlikely to ever need to disambiguate between; in a particular context, this filter will then play a clear role, but one which is hard to determine from the overall shape of the parameters.

\section{Conclusion}

In this work, we investigated using convolutional networks to capture semantic similarity between source documents and potential entity link targets. Using multiple granularities of convolutions to evaluate the compatibility of a mention in context and several potential link targets gives strong performance on its own; moreover, such features also improve a pre-existing entity linking system based on sparse indicator features, showing that these sources of information are complementary.

\section*{Acknowledgments}

This work was partially supported by NSF Grant CNS-1237265 and a Google Faculty Research Award. Thanks to the anonymous reviewers for their helpful comments.

\bibliography{naaclhlt2016}

\begin{thebibliography}{}

\bibitem[\protect\citename{Bentivogli \bgroup et al.\egroup
  }2010]{BentivogliEtAl2010}
Luisa Bentivogli, Pamela Forner, Claudio Giuliano, Alessandro Marchetti,
  Emanuele Pianta, and Kateryna Tymoshenko.
\newblock 2010.
\newblock {Extending English ACE 2005 Corpus Annotation with Ground-truth Links
  to Wikipedia}.
\newblock In {\em Proceedings of the Workshop on {The People's Web Meets NLP:
  Collaboratively Constructed Semantic Resources}}.

\bibitem[\protect\citename{Cheng and Roth}2013]{ChengRoth2013}
Xiao Cheng and Dan Roth.
\newblock 2013.
\newblock {Relational Inference for Wikification}.
\newblock In {\em Proceedings of the Conference on Empirical Methods in Natural
  Language Processing (EMNLP)}.

\bibitem[\protect\citename{Cucerzan}2007]{Cucerzan2007}
Silviu Cucerzan.
\newblock 2007.
\newblock {Large-Scale Named Entity Disambiguation Based on Wikipedia Data}.
\newblock In {\em Proceedings of the Joint Conference on Empirical Methods in
  Natural Language Processing and Computational Natural Language Learning
  (EMNLP-CoNLL)}.

\bibitem[\protect\citename{Dong \bgroup et al.\egroup }2015]{DongEtAl2015}
Li~Dong, Furu Wei, Ming Zhou, and Ke~Xu.
\newblock 2015.
\newblock {Question Answering over Freebase with Multi-Column Convolutional
  Neural Networks}.
\newblock In {\em Proceedings of the 53rd Annual Meeting of the Association for
  Computational Linguistics (ACL) and the 7th International Joint Conference on
  Natural Language Processing (Volume 1: Long Papers)}.

\bibitem[\protect\citename{Durrett and Klein}2014]{DurrettJoint2014}
Greg Durrett and Dan Klein.
\newblock 2014.
\newblock {A Joint Model for Entity Analysis: Coreference, Typing, and
  Linking}.
\newblock In {\em Transactions of the Association for Computational Linguistics
  (TACL)}.

\bibitem[\protect\citename{Fahrni and Strube}2014]{FahrniLatent2014}
Angela Fahrni and Michael Strube.
\newblock 2014.
\newblock A latent variable model for discourse-aware concept and entity
  disambiguation.
\newblock In Gosse Bouma and Yannick~Parmentier 0001, editors, {\em Proceedings
  of the 14th Conference of the European Chapter of the Association for
  Computational Linguistics, EACL 2014, April 26-30, 2014, Gothenburg, Sweden},
  pages 491--500. The Association for Computer Linguistics.

\bibitem[\protect\citename{He \bgroup et al.\egroup }2013]{HeStacking2013}
Zhengyan He, Shujie Liu, Yang Song, Mu~Li, Ming Zhou, and Houfeng Wang.
\newblock 2013.
\newblock Efficient collective entity linking with stacking.
\newblock In {\em EMNLP}, pages 426--435. ACL.

\bibitem[\protect\citename{Heath and Bizer}2011]{HeathWPLink2011}
Tom Heath and Christian Bizer.
\newblock 2011.
\newblock {\em Linked Data: Evolving the Web into a Global Data Space}.
\newblock Morgan \& Claypool, 1st edition.

\bibitem[\protect\citename{Hoffart \bgroup et al.\egroup
  }2011]{HoffartCoNLL2011}
Johannes Hoffart, Mohamed~Amir Yosef, Ilaria Bordino, Hagen F\"{u}rstenau,
  Manfred Pinkal, Marc Spaniol, Bilyana Taneva, Stefan Thater, and Gerhard
  Weikum.
\newblock 2011.
\newblock {Robust Disambiguation of Named Entities in Text}.
\newblock In {\em Proceedings of the Conference on Empirical Methods in Natural
  Language Processing (EMNLP)}.

\bibitem[\protect\citename{Iyyer \bgroup et al.\egroup }2015]{IyyerEtAl2015}
Mohit Iyyer, Varun Manjunatha, Jordan Boyd-Graber, and Hal {Daum\'{e} III}.
\newblock 2015.
\newblock {Deep Unordered Composition Rivals Syntactic Methods for Text
  Classification}.
\newblock In {\em Proceedings of the Association for Computational Linguistics
  (ACL)}.

\bibitem[\protect\citename{Ji and Grishman}2011]{JiGrishman2011}
Heng Ji and Ralph Grishman.
\newblock 2011.
\newblock {Knowledge Base Population: Successful Approaches and Challenges}.
\newblock In {\em Proceedings of the Association for Computational Linguistics
  (ACL)}.

\bibitem[\protect\citename{Kalchbrenner \bgroup et al.\egroup
  }2014]{KalchbrennerEtAl2014}
Nal Kalchbrenner, Edward Grefenstette, and Phil Blunsom.
\newblock 2014.
\newblock A convolutional neural network for modelling sentences.
\newblock In {\em Proceedings of the 52nd Annual Meeting of the Association for
  Computational Linguistics (Volume 1: Long Papers)}, pages 655--665,
  Baltimore, Maryland, June. Association for Computational Linguistics.

\bibitem[\protect\citename{Kim}2014]{KimCNNSent2014}
Yoon Kim.
\newblock 2014.
\newblock {Convolutional Neural Networks for Sentence Classification}.
\newblock In {\em Proceedings of the Conference on Empirical Methods in Natural
  Language Processing (EMNLP)}.

\bibitem[\protect\citename{Levy and Goldberg}2014]{LevyGoldberg2014}
Omer Levy and Yoav Goldberg.
\newblock 2014.
\newblock {Dependency-Based Word Embeddings}.
\newblock In {\em Proceedings of the 52nd Annual Meeting of the Association for
  Computational Linguistics (ACL)}.

\bibitem[\protect\citename{Mikolov \bgroup et al.\egroup }2013]{MikolovW2V2013}
Tomas Mikolov, Ilya Sutskever, Kai Chen, Greg~S Corrado, and Jeff Dean.
\newblock 2013.
\newblock {Distributed Representations of Words and Phrases and their
  Compositionality}.
\newblock In {\em Advances in Neural Information Processing Systems (NIPS) 26},
  pages 3111--3119.

\bibitem[\protect\citename{Milne and Witten}2008]{MilneWitten2008}
David Milne and Ian~H. Witten.
\newblock 2008.
\newblock {Learning to Link with Wikipedia}.
\newblock In {\em Proceedings of the Conference on Information and Knowledge
  Management (CIKM)}.

\bibitem[\protect\citename{Nguyen \bgroup et al.\egroup
  }2014]{NguyenAidaLight2015}
Dat~Ba Nguyen, Johannes Hoffart, Martin Theobald, and Gerhard Weikum.
\newblock 2014.
\newblock {AIDA-light: High-Throughput Named-Entity Disambiguation}.
\newblock In {\em Proceedings of the Workshop on Linked Data on the Web
  co-located with the 23rd International World Wide Web Conference {(WWW)}}.

\bibitem[\protect\citename{{NIST}}2005]{Ace2005}
{NIST}.
\newblock 2005.
\newblock {The ACE 2005 Evaluation Plan}.
\newblock In {\em NIST}.

\bibitem[\protect\citename{Ratinov \bgroup et al.\egroup
  }2011]{RatinovWiki2011}
Lev Ratinov, Dan Roth, Doug Downey, and Mike Anderson.
\newblock 2011.
\newblock {Local and Global Algorithms for Disambiguation to Wikipedia}.
\newblock In {\em Proceedings of the 49th Annual Meeting of the Association for
  Computational Linguistics (ACL)}, pages 1375--1384.

\bibitem[\protect\citename{Shen \bgroup et al.\egroup }2014]{ShenEtAl2014}
Yelong Shen, Xiaodong He, Jianfeng Gao, Li~Deng, and Gr{\'e}goire Mesnil.
\newblock 2014.
\newblock {Learning Semantic Representations Using Convolutional Neural
  Networks for Web Search}.
\newblock In {\em Proceedings of the 23rd International Conference on World
  Wide Web (WWW)}.

\bibitem[\protect\citename{Sun \bgroup et al.\egroup }2015]{SunNNetEntity2015}
Yaming Sun, Lei Lin, Duyu Tang, Nan Yang, Zhenzhou Ji, and Xiaolong Wang.
\newblock 2015.
\newblock {Modeling Mention, Context and Entity with Neural Networks for Entity
  Disambiguation}.
\newblock In {\em Proceedings of the International Joint Conference on
  Artificial Intelligence (IJCAI)}, pages 1333--1339.

\bibitem[\protect\citename{Zeiler}2012]{ZeilerAdadelta2012}
Matthew~D. Zeiler.
\newblock 2012.
\newblock {AdaDelta: An Adaptive Learning Rate Method}.
\newblock {\em CoRR}, abs/1212.5701.

\end{thebibliography}
\bibliographystyle{naaclhlt2016}

\end{document}